\documentclass{article}
\usepackage{graphicx} 
\usepackage{amsmath}
\usepackage{amssymb}
\usepackage{algorithm}
\usepackage{algpseudocode}
\usepackage{multirow}
\usepackage{blindtext}

\title{Enhancing Dialogue Systems with Discourse-Level Understanding Using Deep Canonical Correlation Analysis}
\author{Akanksha Mehndiratta, Krishna Asawa }

\begin{document}

\maketitle
\begin{abstract}
    The evolution of conversational agents has been driven by the need for more contextually aware systems that can effectively manage dialogue over extended interactions. To address the limitations of existing models in capturing and utilizing long-term conversational history, we propose a novel framework that integrates Deep Canonical Correlation Analysis (DCCA) for discourse-level understanding. This framework learns discourse tokens to capture relationships between utterances and their surrounding context, enabling a better understanding of long-term dependencies. Experiments on the Ubuntu Dialogue Corpus demonstrate significant enhancement in response selection, based on the improved automatic evaluation metric scores. The results highlight the potential of DCCA in improving dialogue systems by allowing them to filter out irrelevant context and retain critical discourse information for more accurate response retrieval.

\end{abstract}
\section{Introduction}
Dialogue systems, such as chatbots or virtual assistants, have made substantial progress in generating contextually appropriate responses. However, these systems face a persistent challenge in maintaining coherence and relevance across multiple turns in longer conversations. This is especially difficult when the context becomes complex, with numerous topics, nuanced references, or shifting conversational goals. With the objective of enhanced language modeling, such models often struggle to effectively utilize the entire discourse history, leading to responses that may be locally appropriate but globally inconsistent or irrelevant \cite{mehndiratta2021non}

The core issue is how dialogue systems manage and interpret discourse history. Current models typically rely on the immediate context (e.g., the last few utterances) to generate responses, which can lead to a loss of important information from earlier in the conversation. This limitation becomes more pronounced in longer dialogues, where the context is spread across many turns and may involve intricate dependencies between utterances.

To address this challenge, this study presents a new framework that leverages Deep Canonical Correlation Analysis (DCCA). DCCA is a technique for modeling relationships between two sets of variables by projecting them into a shared latent space. In this context, DCCA is applied to model the relationship between an utterance and its surrounding turns (i.e., the discourse history).
The key idea is to create a shared latent representation that captures the essential elements of the conversation while filtering out irrelevant or redundant information. This allows the model to build a discourse-level understanding of the conversation, enabling it to focus on the most relevant context and select coherent and contextually appropriate responses.

This proposed framework enhances coherence by modeling the relationship between an utterance and its discourse history, ensuring a consistent conversational thread even in extended dialogues\cite{mehndiratta2025modeling}. It also improves relevance filtering by leveraging a shared latent space to prioritize essential context while discarding irrelevant information, resulting in more focused responses. Furthermore, the method is highly scalable, effectively managing complex and varied contexts, making it well-suited for longer conversations.

The proposed framework is evaluated on the Ubuntu Dialogue Corpus, a widely used benchmark in dialogue system research. This corpus consists of multi-turn technical support conversations, making it an ideal testbed for evaluating the ability to maintain coherence and relevance over extended dialogues. The experiments demonstrate that the framework achieves significant improvements in automatic evaluation metrics for response selection tasks. 

\section{Related Work}
Retrieval-based dialog systems maintain conversations by choosing the most appropriate response from a set of candidates, using the conversation history and the user's latest input as context \cite{wu-etal-2017-sequential,zhou2016multi,zhou2018multi,henderson-etal-2019-training}. Typically, these systems encode both the conversation and  the potential responses as vectors within a shared semantic space and use a classifier to evaluate how well they match\cite{lowe2015ubuntu,wu-etal-2017-sequential,zhang2018modeling}. Data-driven dialogue systems generally follow two main approaches: (1) combining all context utterances into a single input \cite{lowe2015ubuntu,Lowe2017TrainingED}, or (2) processing each utterance individually and then aggregating the results\cite{wu-etal-2017-sequential,zhou2018multi,tao2019one}. Among these methods, RNN-based architectures are commonly used for response selection\cite{lowe2015ubuntu,Lowe2017TrainingED,zhou2016multi,wang-jiang-2016-learning}. Early single-turn models, such as RNN, CNN, LSTM, BiLSTM, MV-LSTM, and Match-LSTM, treated the entire context as a single document to compute matching scores with candidate responses. More recent approaches focus on multi-turn matching between context and response\cite{wu-etal-2017-sequential,zhou2018multi}. To better capture interactive information, attention mechanisms have been employed to model relationships across multiple utterances. For instance, Zhang et al.\cite{zhang2018modeling} used self-matching attention to extract key information from each utterance, while Zhou et al. \cite{zhou2018multi} applied stacked self-attention and cross-attention to represent utterances in a response-aware manner. Tao et al. \cite{tao2019multi,tao2019one} enhanced context-to-response matching by stacking multiple interaction blocks. Recent advancements aim to capture matched information at varying granularities, achieving state-of-the-art performance in multi-turn response selection\cite{gu2019interactive,tao2019multi,yuan2019multi}.

In parallel, deep contextualized pre-trained language models (PTMs) have revolutionized NLP by learning universal language representations, significantly improving performance across various tasks. Notable examples include ELMo, GPT, BERT, RoBERTa, XLNet, ALBERT, and ELECTRA. These models differ in their approaches to masking strategies, knowledge injection, language modeling objectives, and efficiency optimization. Despite their success, the potential of PTMs remains underexplored, particularly in adapting them to dialogue systems\cite{gururangan-etal-2020-dont,sun2019fine}. Limited research has focused on tailoring PTMs to the unique challenges of conversational AI, leaving significant room for innovation in this area.

\section{Deep Canonical Correlation Analysis (DCCA)}
Deep Canonical Correlation Analysis (DCCA)\cite{andrew2013deep} is a powerful technique that aims to learn a shared representation of two sets of variables by maximizing their correlation in a common latent space. Traditionally used in the domain of unsupervised learning, DCCA can be extended to handle multi-modal data, aligning features from different domains. By learning representations that capture the relationships between two sets of inputs, DCCA helps uncover hidden dependencies between variables, which is especially valuable in dialogue systems\cite{gao2020cross,Sun2019MultimodalSA}.

With an intent of performing discourse understanding, this study employs DCCA to learn a shared latent space between the utterance and its surrounding context (previous and subsequent turn). This shared representation allows the model to better understand the relationship between an utterance and the broader dialogue history, thereby improving the relevance of the selected response.

The key benefit of DCCA in dialogue systems is its ability to map an utterance and its contextual surroundings into a space where their correlations can be maximized. This ensures that the model can distinguish between relevant and irrelevant knowledge from a given context, significantly improving its response selection capabilities.

\subsection{Feature Extraction Using Deep Canonical Correlation Analysis}
In the traditional design of conversational models, the hidden states generated by recurrent neural networks (RNNs) or transformers are often used to represent the contextual information of a dialogue. However, these hidden states alone may not effectively capture all the inter-relationships that are crucial for understanding discourse at a deeper level. By applying Canonical Correlation Analysis (CCA) to the hidden states, we can extract features that better represent the interdependencies between an utterance and its dialogue history\cite{mehndiratta2024discovering}.

In our approach, we utilize DCCA to align the hidden states from different turns in the conversation, ensuring that the model captures the essential dependencies between the utterance and its surrounding context. The interpretation of these hidden states, combined with the feature extraction process enabled by DCCA, allows the model to maintain a rich, nuanced understanding of the ongoing discourse. This enhances the model’s ability to select more contextually appropriate responses by focusing on the most relevant aspects of the conversation.

By embedding the hidden state vectors from multiple turns of dialogue into a shared subspace, the model not only ensures a more coherent flow of information but also allows the system to filter out irrelevant information. The learned hidden state vectors can then be used to improve the selection of responses, leading to higher-quality dialogue interactions.

The framework to perform feature extraction using DCCA consists of three main components: the encoder, the discourse-level understanding module, and the response selection module.
\begin{enumerate}
    \item Encoder: The encoder is responsible for processing the input dialogue. It consists of a pre-trained language model that encodes both the current utterance and its surrounding utterances into a representation. These representations capture the meaning of words in an utterance.
    
    \item Discourse-Level Understanding Module: This module applies DCCA to learn a shared representation of the current utterance and its context. The DCCA framework maximizes the correlation between the hidden states of different turns, ensuring that the model captures the relationship between an utterance and its preceding and succeeding turns. The discourse-level understanding module generates discourse tokens, which serve as refined features that highlight the most relevant contextual information.

    \item Response Selection Module: Using the discourse tokens generated by the DCCA framework, the response selection module evaluates candidate responses based on their relevance to the conversation. This module employs a ranking mechanism (a similarity measure) to identify the most appropriate response from a set of candidates.
    \end{enumerate}

\section{Dataset}
The Ubuntu Dialogue Corpus\cite{lowe2015ubuntu} contains nearly one million two-person text-based conversations from Ubuntu chat logs, focused on technical support. Each conversation averages eight turns, with a minimum of three. It is comparable in scale to other problem-solving datasets but includes more turns and longer messages. This makes it one of the largest datasets for dialogue system research, featuring human-human, two-way conversations with a large volume of multi-turn interactions.

\subsection{Data Preparation}
To maintain standardization, the Ubuntu dataset is processed to extract (context, response, flag) triples from each dialogue. The flag is a Boolean value indicating whether the response is the actual next utterance in the conversation. The response is the expected outcome that the model aims to identify correctly, while the context includes all previous utterances in the dialogue. Each dialogue generates two triples: one with the correct response (actual next utterance) and another with a false response, randomly chosen. The flag is set to 1 for the correct response and 0 for the false one. To increase difficulty, the task can include multiple incorrect responses instead of just one. This study performs experiments for the following three cases— one incorrect, two incorrect and five incorrect responses.

\section{System Overview}
 The model defines a context C and candidate responses R as input. Here C = {Utt\textsubscript{1}, Utt\textsubscript{2}, . . . , Utt\textsubscript{i}}; where i is the number of utterances in the input context of a conversation; Let R =  = (r\textsubscript{1},r\textsubscript{2} ... r\textsubscript{a}) be the candidate response set. The model aims to determine a score for each candidate r\textsubscript{a} by utilizing the discourse knowledge constructed by performing feature extraction to filter representations relevant to context C.

\subsection{Encoder} 
The encoder processes the input context to generate utterance-level representations. Each input utterance is mapped to a p-dimensional feature vector space using a pre-trained language model. Consequently, each utterance is represented $\in$ $\mathbb{R}$\textsuperscript{n x p}; Here n denotes the number of word tokens in the utterance. These representations effectively capture the semantic meaning of words within the utterance.
    
\subsection{Discourse-Level Understanding Module} 
This module utilizes DCCA to create a shared representation of the current utterance and its surrounding context. By maximizing the correlation between the hidden states of various turns, the DCCA framework ensures that the model effectively captures the connections between an utterance and its adjacent turns, both before and after. The discourse-level understanding module produces discourse tokens, which act as enhanced features that emphasize the most pertinent contextual details.

Suppose an utterance is composed of g word tokens {w\textsubscript{1}, w\textsubscript{2}, $...$, w\textsubscript{g}}. For the initial utterance-pair representation (Utt\textsubscript{i}, Utt\textsubscript{j}) with dimensions g x p and h x p respectively, DCCA constructs a univariate random variable taking a linear combination of its components:$\sum_{j=1}^{p}  \alpha$\textsubscript{j}f\textsubscript{j}. Here f\textsubscript{j} represents the j\textsuperscript{th} feature in the p dimensional subspace. It then calculates the pairwise correlation coefficients for each variable pair to determine the vectors $\alpha$\textsubscript{i} that maximize the total sum of these correlations. As a result, DCCA builds a projection matrix $\Lambda$, which maps the k most correlated word tokens, where k = min(g, h), using the following equation.
\begin{equation}
\Lambda\textsubscript{1}, \Lambda\textsubscript{2} = DCCA(Utt\textsubscript{1}, Utt\textsubscript{2})
\end{equation}
The projections, $\Lambda$\textsubscript{1} and $\Lambda$\textsubscript{2}, consist of word representations that redundantly estimate the same hidden state H, 
The initial k projections derived from DCCA are termed intentions, and unique intentions are aggregated as discourse tokens. These discourse tokens encapsulate critical information about the underlying relationships and hidden structures within the utterance pair, capturing key contextual details relevant to the ongoing dialogue.

Applying DCCA to the established discourse and subsequent utterance acquires additional intents and discourse tokens. Therefore, Examining the connections between discourse-level information introduced at the start of the conversation and the utterances that follow later. The discourse tokens generated by the DCCA-based module ensure that the model filters out irrelevant context and focuses on the critical elements that are most likely to lead to the best response.

\subsection{Response Selection Module}
Each potential response is encoded using a pre-trained language model to produce an utterance-level representation. A matching score is calculated by determining the cosine similarity between this representation and the discourse tokens extracted after analyzing the entire conversation, i.e. context. The candidate responses are then ranked according to their matching scores and the model chooses the most suitable response from the top-ranked options.

\section{Experimental Setting}
\subsection{Model}
DCCA is an efficient tool for extracting features by finding relationships between two sets of variables. As discussed in the previous section, the features extracted by DCCA between an utterance pair are called discourse tokens. These tokens are then utilized to construct discourse-level knowledge using the algorithm \ref{alg:alg2}. The algorithms follow the architectural framework outlined in the previous section to model conversational history. The utterance-level representations created for each utterance in the context are used to derive intents through CCA. 
DCCA is implemented in MVLearn open-source Python library. Feature extraction using DCCA involves methods such as DCCA(NumOfComponents) and dcca.fit(Utt\textsubscript{1}\textsuperscript{T}, Utt\textsubscript{2}\textsuperscript{T}) to fit the model to the input data and generate projections (latent variable pairs) $\Lambda$\textsubscript{Utt1}, $\Lambda$ \textsubscript{Utt2}, each with a shape of (NumOfFeatures, NumOfComponents). The method dcca.transform(Utt\textsubscript{1}\textsuperscript{T}, Utt\textsubscript{2}\textsuperscript{T}) transforms the input utterance-level representations by maximizing their correlation. Intentions (DLU) are derived using the projections obtained from both utterance-level representations. Here, NumOfFeatures corresponds to the number of features that represent word tokens in an utterance (i.e., p), while the number of components is restricted to NumOfComponents = min(number of word tokens in Utt\textsubscript{1}, number of word tokens in Utt\textsubscript{2}).
\begin{algorithm}
\caption{Algorithm for feature extraction using DCCA to determine discourse tokens}
\label{alg:alg2}
\begin{algorithmic}[1]
\State DLU = $\gets$ Utt\textsubscript{1}
\State N $\gets$ 2
\While{$N \neq size$}
\State NumOfProjections = min(len(DLU), len(Utt\textsubscript{N}))
\State dcca = DCCA(NumOfComponents=NumOfProjections)
\State Fit(DLU, Utt\textsubscript{N}) 
\State $\Lambda$\textsubscript{1}, $\Lambda$\textsubscript{2} = Transform(DLU, Utt\textsubscript{N})
\State DLU = Unique($\Lambda$\textsubscript{1}, $\Lambda$\textsubscript{2}, DLU)
\State N $gets$ N + 1
\EndWhile
\end{algorithmic}
\end{algorithm}
\subsection{Baselines}
Several models have been introduced to improve multi-turn response selection. Building on the work of \cite{mehndiratta2025modeling}, which utilizes CCA for feature extraction to identify intents and discourse tokens, this study extends that approach.  In particular, it enhances response coherence by learning discourse tokens with higher correlations, thereby improving discourse-level understanding.

\subsection{Result and Analysis}
The performance of all baselines and our proposed model using Recall is shown in Table \ref{tab:tab1}. It is evident from the table that the model significantly outperforms all other models in terms of most of the metrics on the Ubuntu datasets. 

The insights derived from qualitative metrics reflect the quality of the top-k selected responses, as presented in Table \ref{tab:tab123}. A lower perplexity score indicates that the model more accurately predicts the data, demonstrating a better understanding of the language. The BLEU and Rouge scores range from 0 to 1, with a score of 1 representing a perfect match between the generated text and the reference text. As shown in Table \ref{tab:tab123}, the proposed model achieves higher translation quality and greater diversity in distinct unigrams and bigrams. The results clearly demonstrate that the model excels at retrieving coherent candidate responses and exhibits a stronger understanding of contextual information.

\begin{table}
\caption{Results of the models on the experiment using Recall\textsubscript{n}@k}
    \label{tab:tab1}
    \centering
    \begin{tabular}{|c|c|c|c|c|} 
       \hline
 Model & Recall\textsubscript{all}@20 & Recall\textsubscript{all}@10 & Recall\textsubscript{all}@5 & Recall\textsubscript{all}@3 \\ \hline
 CCA+Global\cite{mehndiratta2025modeling} & 69 & 62 & 57 & 57 \\ \hline
 CCA+Local\cite{mehndiratta2025modeling} & 71 & 64 & 60 & 59 \\  \hline
 MVDF & 73 & 66 & 60 & 60 \\ \hline
 DCCA &  76 & 68 & 65 & 63 \\ \hline
    \end{tabular}
    
\end{table}

\begin{table}
\caption{Results of the models on various popular evaluation metrics in NLP}
    \label{tab:tab123}
    \centering
    \begin{tabular}{|p{2.3cm}|p{1.8cm}|p{1.5cm}|p{1.5cm}|p{2cm}|p{2cm}|p{2cm}|} 
       \hline
 Model & Perplexity & BLEU Score & Rouge - 1 Score & Rouge - L Score & Distinct N Unigram & Distinct N Bigram\\ \hline
 CCA+Global\cite{mehndiratta2025modeling} & 23.1891 & 0.3345 & 0.1469 & 0.1365 & 0.1800 & 0.4554\\ \hline
 CCA+Local\cite{mehndiratta2025modeling} & 28.3874 & 0.3192 & 0.1476 & 0.1374 & 0.1774 & 0.4490\\  \hline
 MVDF & 18.012 & 0.3921 & 0.1834 & 0.1732 & 0.1853 & 0.4657 \\ \hline
 DCCA & 25.7834 & 0.3211 & 0.1835 & 0.1742 & 0.1802 & 0.4512 \\ \hline
    \end{tabular}
    
\end{table}

\section{Conclusion}
In this paper, we introduced a discourse framework based on Deep Canonical Correlation Analysis (DCCA) to improve dialogue systems' understanding of conversational context. Our model achieves a better representation of dialogue history by learning discourse tokens that capture the relationships between an utterance and its surrounding turns. This allows the system to filter out irrelevant context and focus on the most important information for response selection. Our experimental results on the Ubuntu Dialogue Corpus demonstrate significant improvements in automatic evaluation metrics, highlighting the effectiveness of this approach in real-world dialogue systems. 

The success of this approach carries significant implications for dialogue systems, particularly in improving long-term context management. By effectively handling long-term dependencies, the framework becomes highly valuable for applications such as customer support, virtual assistants, and social chatbots, where maintaining context over extended interactions is crucial. Additionally, by filtering out irrelevant information, the system enhances response precision and relevance, ultimately leading to a more satisfying user experience. Furthermore, the use of Deep Canonical Correlation Analysis (DCCA) to model discourse-level relationships lays a strong foundation for future research, enabling the development of more sophisticated dialogue models capable of managing complex conversational dynamics. A future direction could be to expand this framework by adding features like speaker intent and sentiment.


\begin{thebibliography}{99}
	
	\bibitem{andrew2013deep}
	G.~Andrew, R.~Arora, J.~Bilmes, and K.~Livescu.
	\newblock Deep canonical correlation analysis.
	\newblock In {\em International conference on machine learning}, pages 1247--1255. PMLR, 2013.
	
	\bibitem{gao2020cross}
	Q.~Gao, H.~Lian, Q.~Wang, and G.~Sun.
	\newblock Cross-modal subspace clustering via deep canonical correlation analysis.
	\newblock In {\em Proceedings of the AAAI Conference on artificial intelligence}, volume~34, pages 3938--3945, 2020.
	
	\bibitem{gu2019interactive}
	J.-C. Gu, Z.-H. Ling, and Q.~Liu.
	\newblock Interactive matching network for multi-turn response selection in retrieval-based chatbots.
	\newblock In {\em Proceedings of the 28th ACM International Conference on Information and Knowledge Management}, CIKM '19, page 2321â€“2324, New York, NY, USA, 2019. Association for Computing Machinery.
	
	\bibitem{gururangan-etal-2020-dont}
	S.~Gururangan, A.~Marasovi{\'c}, S.~Swayamdipta, K.~Lo, I.~Beltagy, D.~Downey, and N.~A. Smith.
	\newblock Don`t stop pretraining: Adapt language models to domains and tasks.
	\newblock In D.~Jurafsky, J.~Chai, N.~Schluter, and J.~Tetreault, editors, {\em Proceedings of the 58th Annual Meeting of the Association for Computational Linguistics}, pages 8342--8360, Online, July 2020. Association for Computational Linguistics.
	
	\bibitem{henderson-etal-2019-training}
	M.~Henderson, I.~Vuli{\'c}, D.~Gerz, I.~Casanueva, P.~Budzianowski, S.~Coope, G.~Spithourakis, T.-H. Wen, N.~Mrk{\v{s}}i{\'c}, and P.-H. Su.
	\newblock Training neural response selection for task-oriented dialogue systems.
	\newblock In A.~Korhonen, D.~Traum, and L.~M{\`a}rquez, editors, {\em Proceedings of the 57th Annual Meeting of the Association for Computational Linguistics}, pages 5392--5404, Florence, Italy, July 2019. Association for Computational Linguistics.
	
	\bibitem{Lowe2017TrainingED}
	R.~Lowe, N.~Pow, I.~Serban, L.~Charlin, C.-W. Liu, and J.~Pineau.
	\newblock Training end-to-end dialogue systems with the ubuntu dialogue corpus.
	\newblock {\em Dialogue Discourse}, 8:31--65, 2017.
	
	\bibitem{lowe2015ubuntu}
	R.~Lowe, N.~Pow, I.~Serban, and J.~Pineau.
	\newblock The ubuntu dialogue corpus: A large dataset for research in unstructured multi-turn dialogue systems.
	\newblock {\em arXiv preprint arXiv:1506.08909}, 2015.
	
	\bibitem{mehndiratta2021non}
	A.~Mehndiratta and K.~Asawa.
	\newblock Non-goal oriented dialogue agents: state of the art, dataset, and evaluation.
	\newblock {\em Artificial Intelligence Review}, 54(1):329--357, 2021.
	
	\bibitem{mehndiratta2024discovering}
	A.~Mehndiratta and K.~Asawa.
	\newblock Discovering elementary discourse units in textual data using canonical correlation analysis.
	\newblock {\em International Journal of Performability Engineering}, 20(12), 2024.
	
	\bibitem{mehndiratta2025modeling}
	A.~Mehndiratta and K.~Asawa.
	\newblock Modeling discourse for dialogue systems using spectral learning.
	\newblock {\em International Journal of Performability Engineering}, 21(2):65, 2025.
	
	\bibitem{sun2019fine}
	C.~Sun, X.~Qiu, Y.~Xu, and X.~Huang.
	\newblock How to fine-tune bert for text classification?
	\newblock In {\em Chinese computational linguistics: 18th China national conference, CCL 2019, Kunming, China, October 18--20, 2019, proceedings 18}, pages 194--206. Springer, 2019.
	
	\bibitem{Sun2019MultimodalSA}
	Z.~Sun, P.~K. Sarma, W.~A. Sethares, and E.~P. Bucy.
	\newblock Multi-modal sentiment analysis using deep canonical correlation analysis.
	\newblock In {\em Interspeech}, 2019.
	
	\bibitem{tao2019multi}
	C.~Tao, W.~Wu, C.~Xu, W.~Hu, D.~Zhao, and R.~Yan.
	\newblock Multi-representation fusion network for multi-turn response selection in retrieval-based chatbots.
	\newblock In {\em Proceedings of the twelfth ACM international conference on web search and data mining}, pages 267--275, 2019.
	
	\bibitem{tao2019one}
	C.~Tao, W.~Wu, C.~Xu, W.~Hu, D.~Zhao, and R.~Yan.
	\newblock One time of interaction may not be enough: Go deep with an interaction-over-interaction network for response selection in dialogues.
	\newblock In {\em Proceedings of the 57th annual meeting of the association for computational linguistics}, pages 1--11, 2019.
	
	\bibitem{wang-jiang-2016-learning}
	S.~Wang and J.~Jiang.
	\newblock Learning natural language inference with {LSTM}.
	\newblock In K.~Knight, A.~Nenkova, and O.~Rambow, editors, {\em Proceedings of the 2016 Conference of the North {A}merican Chapter of the Association for Computational Linguistics: Human Language Technologies}, pages 1442--1451, San Diego, California, June 2016. Association for Computational Linguistics.
	
	\bibitem{wu-etal-2017-sequential}
	Y.~Wu, W.~Wu, C.~Xing, M.~Zhou, and Z.~Li.
	\newblock Sequential matching network: A new architecture for multi-turn response selection in retrieval-based chatbots.
	\newblock In R.~Barzilay and M.-Y. Kan, editors, {\em Proceedings of the 55th Annual Meeting of the Association for Computational Linguistics (Volume 1: Long Papers)}, pages 496--505, Vancouver, Canada, July 2017. Association for Computational Linguistics.
	
	\bibitem{yuan2019multi}
	C.~Yuan, W.~Zhou, M.~Li, S.~Lv, F.~Zhu, J.~Han, and S.~Hu.
	\newblock Multi-hop selector network for multi-turn response selection in retrieval-based chatbots.
	\newblock In {\em Proceedings of the 2019 conference on empirical methods in natural language processing and the 9th international joint conference on natural language processing (EMNLP-IJCNLP)}, pages 111--120, 2019.
	
	\bibitem{zhang2018modeling}
	Z.~Zhang, J.~Li, P.~Zhu, H.~Zhao, and G.~Liu.
	\newblock Modeling multi-turn conversation with deep utterance aggregation.
	\newblock In E.~M. Bender, L.~Derczynski, and P.~Isabelle, editors, {\em Proceedings of the 27th International Conference on Computational Linguistics}, pages 3740--3752, Santa Fe, New Mexico, USA, Aug. 2018. Association for Computational Linguistics.
	
	\bibitem{zhou2016multi}
	X.~Zhou, D.~Dong, H.~Wu, S.~Zhao, D.~Yu, H.~Tian, X.~Liu, and R.~Yan.
	\newblock Multi-view response selection for human-computer conversation.
	\newblock In {\em Proceedings of the 2016 conference on empirical methods in natural language processing}, pages 372--381, 2016.
	
	\bibitem{zhou2018multi}
	X.~Zhou, L.~Li, D.~Dong, Y.~Liu, Y.~Chen, W.~X. Zhao, D.~Yu, and H.~Wu.
	\newblock Multi-turn response selection for chatbots with deep attention matching network.
	\newblock In {\em Proceedings of the 56th Annual Meeting of the Association for Computational Linguistics (Volume 1: Long Papers)}, pages 1118--1127, 2018.
	
\end{thebibliography}
\bibliographystyle{unsrt}

\end{document}